\documentclass[10pt,twocolumn,letterpaper]{article}

\usepackage{cvpr}
\usepackage{times}
\usepackage{epsfig}
\usepackage{graphicx}
\usepackage{amsmath}
\usepackage{amssymb}

\usepackage{subfigure}

\usepackage{algorithm}
\usepackage{algorithmic}

\usepackage{placeins}
\usepackage{multirow}
\usepackage{booktabs}

\usepackage[pagebackref=true,breaklinks=true,letterpaper=true,colorlinks,bookmarks=false]{hyperref}

\cvprfinalcopy 

\def\cvprPaperID{2345} 

\ifcvprfinal\fi
\begin{document}

\title{DropFilter: A Novel Regularization Method for Learning Convolutional Neural Networks}

\author{Hengyue Pan\
	National University of Defense Technology\\
	109 Deya Road, Changsha, Hunan, China\\
	{\tt\small hengyuepan@nudt.edu.cn}
	\and
	Hui Jiang\\
	York University\\
	4700 Keele Street, Toronto, Ont., Canada\\
	{\tt\small hj@cse.yorku.ca}
	\and
	Xin Niu\\
	National University of Defense Technology\\
	109 Deya Road, Changsha, Hunan, China\\
	{\tt\small niuxin@nudt.edu.cn}
	\and
	Yong Dou\\
	National University of Defense Technology\\
	109 Deya Road, Changsha, Hunan, China\\
	{\tt\small yongdou@nudt.edu.cn}
}

\maketitle

\begin{abstract}
   The past few years have witnessed the fast development of different regularization methods for deep learning models such as fully-connected deep neural networks (DNNs) and Convolutional Neural Networks (CNNs). Most of previous methods mainly consider to drop features from input data and hidden layers, such as Dropout, Cutout and DropBlocks. DropConnect select to drop connections between fully-connected layers. By randomly discard some features or connections, the above mentioned methods control the overfitting problem and improve the performance of neural networks. In this paper, we proposed two novel regularization methods, namely DropFilter and DropFilter-PLUS, for the learning of CNNs. Different from the previous methods, DropFilter and DropFilter-PLUS selects to modify the convolution filters. For DropFilter-PLUS, we find a suitable way to accelerate the learning process based on theoretical analysis. Experimental results on MNISTshow that using DropFilter and DropFilter-PLUS may improve performance on image classification tasks.
\end{abstract}

\section{Introduction}

In the past few years, deep learning models, such as fully-connected deep neural networks (DNNs) and Convolutional Neural Networks (CNNs), achieve huge success in many different tasks. By learning from large-scale labeled databases end-to-end in a supervised way, deep learning models provide good performance in classification, recognition and so on. However, due to huge number of learnable parameters, the processes of deep models training are always affected by the problem of overfitting. Along with the development of deep learning, several regularization methods with different consideration has been proposed to deal with overfitting. Early stopping \cite{Yao2007On} is a simple way to alleviate overfitting, which tries to stop training in the early stage to make the deep learning models have 'no time' to learn too many noises from training samples. Unsupervised pre-training \cite{Erhan2010Why} is another widely-applied method to deal with overfitting, which introduced layer-wise greedy pre-training before the supervised learning procedure to improve performance. In practice, those regularization methods show their advantages in many tasks.

Starting from Dropout \cite{srivastava2014dropout}, large numbers of 'drop-related' methods are proposed to regularize deep learning models. By randomly dropping a pre-defined portion of hidden nodes in the network during every iteration of training, Dropout introduces random noises into models and in fact enjoys the advantages of model combination. Many experimental results prove that Dropout can significantly reduce overfitting and improve network performance on test sets. Based on Dropout, Standout \cite{ba2013adaptive} was proposed, which introduces a binary belief network to compute the drop rate for every node. \cite{Tompson2014Efficient} argued that using regular Dropout in fully-convolutional networks is not good enough since it has not taken spatial relationships of feature maps into account. Therefore, a generalization of Dropout, called SpatialDropout, was proposed to guarantee that the adjacent pixels across all feature maps should be dropped or kept at the same time. To deal with training problems of very deep CNNs, a regularization method called Stochastic Depth \cite{Huang2016dropblock} was designed. Instead of drop part of nodes in the hidden layers, Stochastic Depth randomly drop a part of layers by using identity function to bypass them in every mini-batch. Residual networks \cite{he2015deep} can be viewed as a special situation of Stochastic Depth. \cite{Golnaz2018DropBlock} hoped to take the continuity of images into account, thus proposed DropBlock, which randomly drop some continuous region in feature maps. Different from the above mentioned random dropping methods, DropWeak \cite{Korchi2018DropWeak} introduced more certainty into the method: it sets all weak weights in the network to zero.

Another kind of drop methods tend to focus on input features. CutOut \cite{Devries2017Cutout} removes a fixed-size patch at a randomly selected location from input images. This simple method achieves good performance on some image databases. Random Erasing \cite{zhong2017random} is similar with Cutout, but the only difference is that it use random value to fill the removed region instead of zero-filling. In \cite{Yang2018DropBand}, DropBand was proposed to regularize input data. Besides RGB channels, DropBand introduces the NIR channel (Near InfraRed) as an extra data channel. Then it trains a CNN for 4 times, and in each time one channel should be droped. This method works well on remote sensing image classification. 

In \cite{wan2013regularization}, DropConnect was proposed to prevent overfitting. Different from all drop-related methods above, DropConnect randomly discard connections between fully-connected layers instead of input features or hidden nodes. DropConnect shows better performance in many databases comparing with Dropout. \cite{Iosifidis2015DropELM} applied both Dropout and DropConnect into Extreme Learning Machine (ELM) and achieved good performance. 

In this paper, we propose two closely related novel regularization method, namely DropFilter and DropFilter-PLUS, to improve performance of CNNs. Instead of dealing with input features, hidden feature maps or connections between fully-connected layers, DropFilter select to randomly drop informations from convolution filters. With the moving of convolution filters on the feature maps, DropFilter-PLUS discards different elements to even increase the uncertainty of the dropping procedure. Experimental results indicate that the two proposed methods can prevent overfitting and obviously improve classification performance on several widely-used image databases. 

\section{Related Works}

Generally speaking, the most important basis of our research includes Dropout and DropConnect. In this section, we review the main idea of Dropout and DropConnect, and show the motivation of the proposed DropFilter. 

\subsection{Dropout}

\cite{srivastava2014dropout} was originally proposed the Dropout method to serve as a overfitting-preventing method for the learning of fully-connected DNNs. \cite{wager2013dropout} shows that Dropout can be viewed as an adaptive regularizer. During the training process, for each DNN layer, every node may be dropped with probability $p$. We use $z^l$ to denote the output vector of the $l$th layer, $W^l$ and $b^l$ the weights and bias respectively, and $f$ the non-linear activation function. Then we have:

\begin{equation}
\label{Eq-regularForward}
z^l = f(W^l z^{l-1} + b^l)
\end{equation}

Based on the basic idea of Dropout, when we apply it into the DNNs, Eq.~\ref{Eq-regularForward} should be revised as:

\begin{equation}
\label{Eq-dropout}
z^l = r \cdot f(W^l z^{l-1} + b^l)
\end{equation}

where $\cdot$ is the element-wise product, and $r$ follow the Bernoulli distribution with probability $1-p$ (which means the Dropout rate is $p$):

\begin{flalign}
\label{Eq-Bernoulli}
\begin{split}
& Pr(r_i = 1) = 1-p, {\text and}\\
& Pr(r_i = 0) = p
\end{split}
\end{flalign}

In the test process, we do not apply Dropout in all layers. Instead, all weights are scaled using the factor $p$. This is used to simulate model combination that merges all possible models generated by Dropout. Experiments show that applying Dropout can obviously increase the test performance of DNNs. Moreover, more researches prove that Dropout also works well in CNNs by randomly drop pixels in feature maps. 

\subsection{DropConnect}

DropConnect \cite{wan2013regularization} can be viewed as a generalization of Dropout. For each pair of adjacent fully-connected layers, DropConnect drops connections between them with probability $p$ rather than drop nodes in hidden layers. DropConnect generates sparse-connected layers, which can be described in Eq.~\ref{Eq-dropconnect}

\begin{equation}
\label{Eq-dropconnect}
z^l = f( (r^w \cdot W^l) z^{l-1} + (r^b \cdot b^l) )
\end{equation}

where $r^w$ and $r^b$ denotes the mask for weights and bias respectively. \cite{wan2013regularization} shows that DropConnect has better performance in many datasets comparing with Dropout. \cite{Smirnov2014Comparison} implies that applying both Dropout and DropConnect results in even better performance on ImageNet database \cite{deng2009imagenet}. 

\subsection{Motivation}

\begin{figure*}[h]
	\begin{center}
		\centerline{\includegraphics[width=1.5\columnwidth]{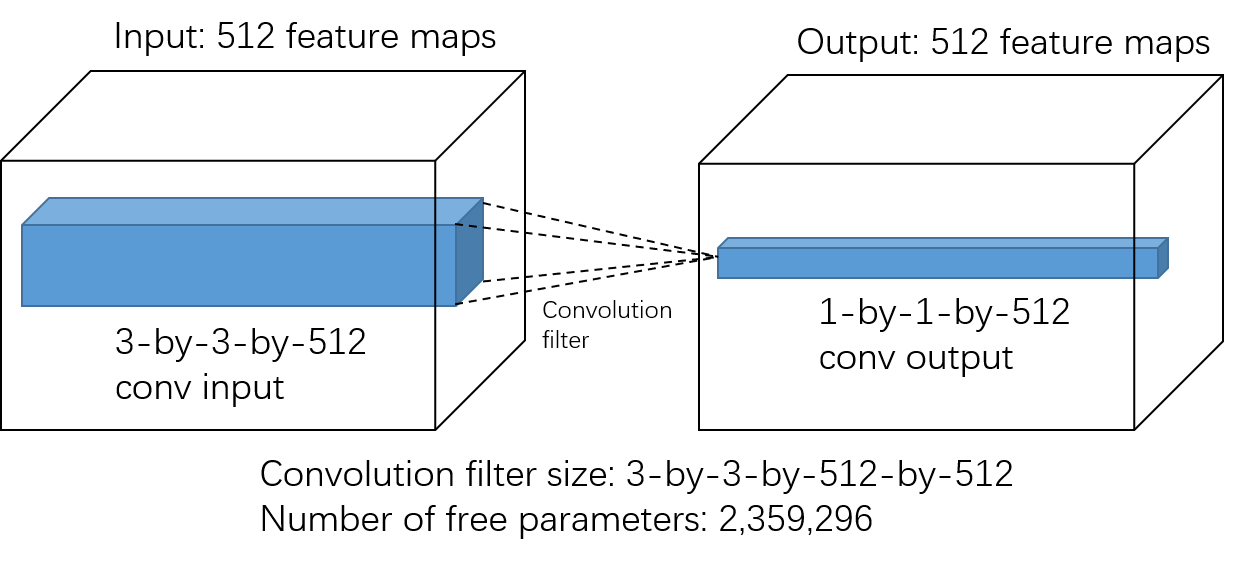}}
		\caption{Illustration of the number of learnable parameters in convolution filters. }
		\label{ConvFilterSize}
	\end{center}
\end{figure*}

In previous researches, Dropout-like methods tend to drop elements from input features or hidden layers, while DropConnect-like methods select to drop connections between fully-connected layers. However, for CNNs, convolution layers are connected using convolution filters, which are not considered in previous regularization methods. Even though the introducing of convolution filters obviously reduce the number of learnable parameters comparing with DNNs, the number of parameters is still very large when CNNs going deep and may still cause overfitting. Figure~\ref{ConvFilterSize} depicts this fact. 

If we try to introduce random feature into convolution filters by randomly drop some elements, we may also prevent overfitting and improve test performance. Therefore, DropFilter is proposed to fulfill this object. The rest of this paper is organized as follows: Section~\ref{DropFilter} provides the details of DropFilter method, and we also present more analysis and implementation details. Section~\ref{Experiments} proposes experimental results on different databases. We conclude our paper in Section~\ref{Conclusion}. 

\section{The Proposed Method}
\label{DropFilter}

In this section, we describe the proposed DropFilter and DropFilter-PLUS method, as well as more detailed information about it. DropFilter can be used for the learning of CNNs as regularizer to prevent overfitting and improve the generalization ability of the network. 


\subsection{DropFilter}

In a convolution layer, the convolution operation in the forward pass can be written as:

\begin{equation}
\label{Eq-convForward}
z^l = f(F^l * z^{l-1} + b^l)
\end{equation}

where $l$ denotes the number of layer, $F$ denotes the convolution filter, and $*$ denotes the convolution operation. Similar with DropConnect, in the forward pass, we randomly generate a $0-1$ mask for each convolution filter and bias in the CNN  at the same time by using Bernoulli distribution. Therefore, DropFilter can be described as below:

\begin{equation}
\label{Eq-DropFilterForward}
z^l = f( (r^F \cdot F^l) * z^{l-1} + (r^b \cdot b^l) )
\end{equation}

In the backward pass, we apply error back-propagation algorithm to calculate the gradients of filters and biases, and use stochastic gradient descent (SGD) to update them. After we get the gradients of error signal with respect to filters and biases, we should use the same masks to post-process them:

\begin{flalign}
\label{Eq-convBackward}
\begin{split}
& F_{i+1}^l = F_i^l - \lambda (r^F \cdot \frac{\partial E}{\partial F_i^l})  \\
& b_{i+1}^l = b_i^l - \lambda (r^b \cdot \frac{\partial E}{\partial b_i^l})  \\ 
\end{split}
\end{flalign}

where $i$ is the number of iteration, $E$ is the error signal, and $\lambda$ is the learning rate. By doing this we guarantee that only the active elements in forward pass can make contribution on filters and biases updates. Algorithm~\ref{alg:DropFilter} describes the detailed procedure of DropFilter. 

\begin{algorithm}[htb]
	\caption{Training CNNs using DropFilter}
	\label{alg:DropFilter}
	\begin{algorithmic}
		\STATE {\bfseries Input:} Learning rate $\lambda$, Filters $F_{i}$ and Biases $b_i$ 
		
		\STATE {\bfseries Output:} Updated Filters $F_{i+1}$ and Biases $b_{i+1}$
		
		\STATE {\bfseries Forward Pass:}
		\STATE Randomly generate masks $r^F$ and $r_b$ follow $Bernoulli(1-p)$ distribution
		\STATE Do forward pass: $z^l = f( (r^F \cdot F_i) * z^{l-1} + (r^b \cdot b_i) )$
		
		\STATE {\bfseries Backward Pass:}
		\STATE Calculate gradients of Error w.r.t. $F_i$ and $b_i$: $\frac{\partial E}{\partial F_i}$ and $\frac{\partial E}{\partial b_i}$
		
		\STATE Using masks in forward pass to post-process gradients: $\frac{\partial E}{\partial F_i} \gets r^F \cdot \frac{\partial E}{\partial F_i}$, $\frac{\partial E}{\partial b_i} \gets r^b \cdot \frac{\partial E}{\partial b_i}$
		
		\STATE Update $F$ and $b$: $F_{i+1} \gets F_{i} - \lambda \frac{\partial E}{\partial F_i} $, $b_{i+1} \gets b_{i} - \lambda \frac{\partial E}{\partial b_i}$
		
	\end{algorithmic}
\end{algorithm}

\subsection{DropFilter-PLUS}

\begin{figure*}[h]
	\begin{center}
		\centerline{\includegraphics[width=1.5\columnwidth]{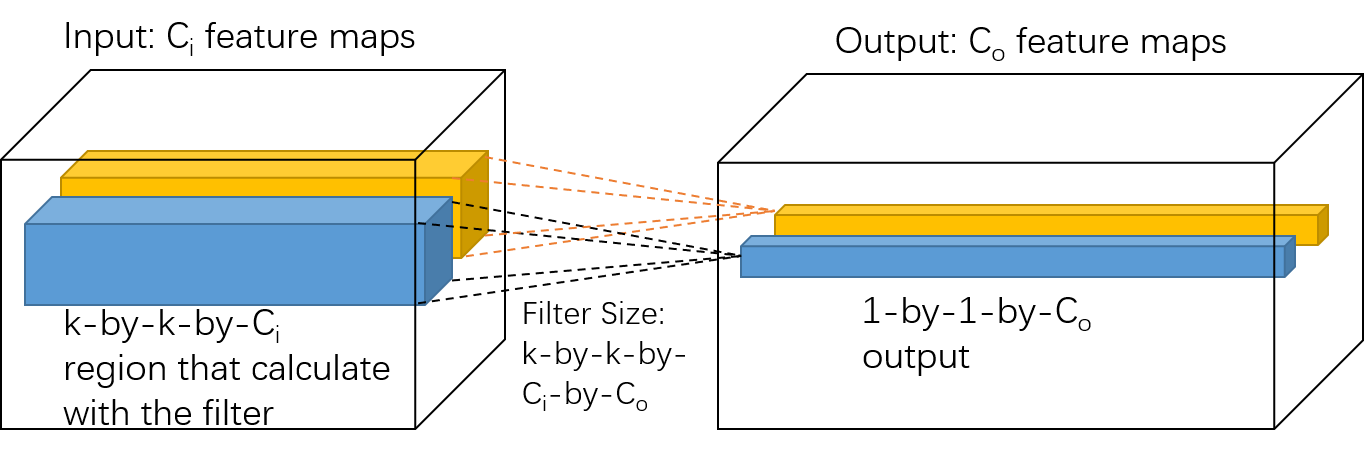}}
		\caption{ The moving procedure in the convolution calculation. }
		\label{ConvMoving}
	\end{center}
\end{figure*}

To even improve the random feature of DropFilter, we define DropFilter-PLUS, which may further decrease the potential problems of overfitting. During the forward pass, we move the convolution filter on input feature maps and calculate every part of output feature maps (see Figure~\ref{ConvMoving} for more details). Along with movings of the filter, we may use different masks to randomly drop elements of the filter. Therefore, different locations on input feature maps may corresponding to different convolution filters. 

Unfortunately, for convolution layers, especially those with large feature maps, we should move convolution filters for many times. For instance, assuming that we have 224-by-224 input feature maps with 64 channels, and 3-by-3-by-64-by-128 convolution filter, it is easy to know that we should move the filter on the input feature maps for $222 * 222 = 49,284$ times, which means we should generate $49,284$ different masks and multiply them with the filter and bias. This may significantly reduce the learning efficiency. 

To accelerate DropFilter-PLUS, we consider to avoid to generate the masks one by one. Based on the definition of convolution operation, we have:

\begin{equation}
\label{Eq-convDef}
a_{x,y}^l = \sum_{x^{\prime}=0}^{k-1} \sum_{y^{\prime}=0}^{k-1} (r_{x^{\prime}, y^{\prime}}^F F_{x^{\prime}, y^{\prime}}^l) z_{x-x^{\prime}, y- y^{\prime}}^{l-1} + r_{x, y}^b b_{x, y}^l
\end{equation}

where $k$ is the convolution filter size, and $a^l$ is the convolution result, and we have $z^l = f(a^l)$. Eq.~\ref{Eq-convDef} implies that the introducing of $0-1$ masks equals to randomly shrink the value of $a^l$ since the operations in Eq.~\ref{Eq-convDef} are linear. Therefore, we may apply a random mask $r^{output}$ that follow the uniform distribution (all elements in the mask should lay between $0$ and $1$) to element-wise multiply with the layer output. Since each column in the output corresponding to different shrink rate, the introduction of $r^{output}$ may simulate the huge number of masks for the filter and bias. 

For the backward pass, we can use the average over all random masks to post-process the gradient of the filter. Assuming that we have $n$ different masks that follow Bernoulli distribution with probability $1-p$, then according to central limit theorem we have:

\begin{equation}
\label{Eq-centrallimit}
\frac{1}{n} \sum_{i=1}^{n} r_i \sim N(1-p, \frac{p(1-p)}{n})
\end{equation}

therefore, in the backward pass we use two new masks $r^{F-norm}$ and $r^{b-norm}$ that follows the normal distribution $N(1-p, \frac{p(1-p)}{n})$ to multiply with gradients of $F$ and $b$. In this way, we obviously accelerate the learning speed of DropFilter-PLUS. Algorithm~\ref{alg:DropFilter} describes the details of DropFilter-PLUS method.

\begin{algorithm}[htb]
	\caption{Training CNNs using DropFilter-PLUS}
	\label{alg:DropFilterPLUS}
	\begin{algorithmic}
		\STATE {\bfseries Input:} Learning rate $\lambda$, Filters $F_{i}$ and Biases $b_i$ 
		
		\STATE {\bfseries Output:} Updated Filters $F_{i+1}$ and Biases $b_{i+1}$
		
		\STATE {\bfseries Forward Pass:}
		\STATE Do forward pass: $z^l = f(  F_i * z^{l-1} + b_i )$
		\STATE Randomly generate mask $r^{output}$ uniform distribution that between $0$ and $1$
		\STATE Update the layer output: $z^l \gets r^{output} \cdot z^l$
		
		\STATE {\bfseries Backward Pass:}
		\STATE Calculate gradients of Error w.r.t. $F_i$ and $b_i$: $\frac{\partial E}{\partial F_i}$ and $\frac{\partial E}{\partial b_i}$
		
		\STATE Generate masks $r^{F-norm}$ and $r^{b-norm}$ follow normal distribution: $N(1-p, \frac{p(1-p)}{n})$
		
		\STATE Using the masks to post-process gradients: $\frac{\partial E}{\partial F_i} \gets r^{F-norm} \cdot \frac{\partial E}{\partial F_i}$, $\frac{\partial E}{\partial b_i} \gets r^{b-norm} \cdot \frac{\partial E}{\partial b_i}$
		
		\STATE Update $F$ and $b$: $F_{i+1} \gets F_{i} - \lambda \frac{\partial E}{\partial F_i} $, $b_{i+1} \gets b_{i} - \lambda \frac{\partial E}{\partial b_i}$
		
	\end{algorithmic}
\end{algorithm}



\section{Experimental Results}
\label{Experiments}

In this section, we apply the proposed DropFilter and DropFilter-PLUS in CNNs for image classification tasks. We use a widely-used database  MNIST to test the classification performance. In our experiments, we compare our proposed methods with several previous regularization methods such as Dropout and DropConnect. We use test error as the performance measurement. 

Our computing platform includes Intel Xeon E5-1650 CPU (6 cores), 64 GB memory and a Nvidia Geforce GTX 1070 GPU (8 GB memory). Our method is implemented with MatConvNet \cite{MatConvNet-2014}, which is a CUDA based CNN toolbox in Matlab.


\subsection{MNIST}

MNIST \cite{lecun1998gradient} is a widely-used hand-written digits recognition database, which contains 60,000 training images and 10,000 test images, and the images are grey-scale and 28-by-28 in size. 

\begin{figure}[htb]
	\begin{center}
		\centerline{\includegraphics[width=0.85\columnwidth]{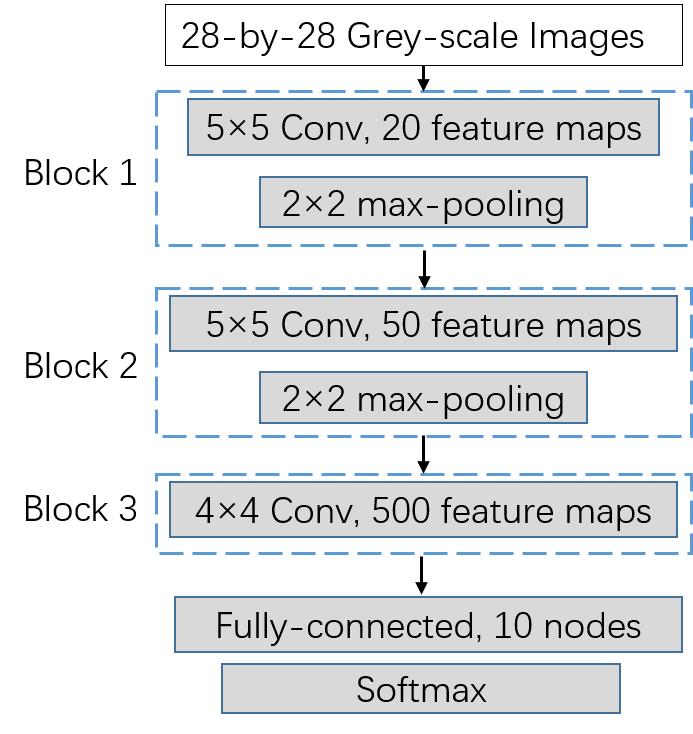}}
		\caption{ The CNN structure of MNIST experiments. }
		\label{MNISTNet}
	\end{center}
\end{figure}

In our MNIST experiments, we apply a CNN with 3 convolutional layers and 1 fully-connected layer. Notice that when used in fully-connected layer, the proposed DropFilter and DropFilter-PLUS degenerate to DropConnect. The whole network structure is shown in Figure~\ref{MNISTNet}. During the experiments, we use 100 batch-size mini-batch SGD to train the network for 30 epochs. The initial learning rate is $0.002$, and we halve it every 2 epochs. For Dropout, DropConnect and the proposed methods, we use $0.5$ as the drop rate. 

We compare the proposed to regularization methods with Dropout and DropConnect to show the advantages of them. Moreover, we also show the performances of the situations that Dropout working with DropFilter or DropFilter-PLUS. Table~\ref{table:MNISTExperiments} shows that DropFilter and DropFilter-PLUS show better performance than Dropout and DropConnect, and Dropout + DropFilter may result in even better performance. 

\begin{table}[ht]
	\caption{Classification error rates on MNIST test set}
	\label{table:MNISTExperiments}
	\begin{center}
		\begin{tabular}{c|c}
			\toprule[2pt]
			Regularization Methods      & Test Error (Top-1)   \\ 
			\midrule[1pt]
			Dropout                     & 0.93\%       \\ 
			DropConnect                 & 0.89\%        \\ 
			DropFilter                  & - \\ 
			DropFilter-PLUS             & 0.82\%  \\ 
			Dropout+DropFilter          & - \\ 
			Dropout+DropFilter-PLUS     & -  \\ 
			\bottomrule[2pt]			
		\end{tabular} 
	\end{center}
\end{table}


\section{Conclusion}
\label{Conclusion}

In this paper, we proposed DropFilter and DropFilter-PLUS to prevent overfitting and improve performance for CNNs. Different from the previous methods like Dropout and DropConnect, the proposed methods select to drop elements from convolution filters, which also has potential to increase the generalization ability. DropFilter randomly discards elements in the convolution filters. By contrast, DropFilter-PLUS introduces even more uncertainty by dropping different elements during the procedure that the convolution filter moving on the input feature maps. Based on the experimental results on MNIST, we can learn that the proposed regularization methods can deal with the problem of overfitting and obviously improve the classification performances. 

{\small
\bibliographystyle{ieee}
\bibliography{dropFilter}
}

\end{document}